\documentclass{article}

\usepackage{arxiv}

\usepackage[utf8]{inputenc} 
\usepackage[T1]{fontenc}    
\usepackage{hyperref}       
\usepackage{url}            
\usepackage{booktabs}       
\usepackage{amsfonts,amsmath,amssymb}       
\usepackage{nicefrac}       
\usepackage{microtype}      
\usepackage{graphicx}
\usepackage{caption}

\usepackage{color, colortbl}

\usepackage{mathptmx,amsthm}
\usepackage{xcolor}
\usepackage{multirow}
\usepackage{subcaption}
\usepackage{algorithm,algorithmic}
\usepackage{textcomp}
\usepackage{float}

\newtheorem{proposition}{Proposition}

\newcommand{\commenttxt}[1]{}

\newcommand{\mybar}{\kern1pt\rule[-\dp\strutbox]{.8pt}{\baselineskip}\kern1pt}

\renewcommand{\mytagline}{} 
\renewcommand{\mydate}{July 10, 2026} 

\theoremstyle{definition}

\title{TVT-PAPD: Pathology-Aware Prototype Distillation for Self-Supervised Whole Slide Image Classification}
\author{
  Ramesh Naidu Laveti\textsuperscript{1},
  Jaya~Sreevalsan-Nair\textsuperscript{1}\thanks{\texttt{jnair@iiitb.ac.in}},
  T K Srikanth
  \\
  \textsuperscript{1}E-Health Research Center,\\
  International Institute of Information Technology Bangalore, Karnataka 560100, India. \\
  \texttt{https://ehrc.iiitb.ac.in/} \\
}

\begin{document}
\maketitle

\begin{abstract}
Self-supervised learning (SSL) has emerged as an effective paradigm for learning transferable representations from large-scale unlabeled whole slide images (WSIs). However, existing SSL methods primarily learn generic visual features and often fail to explicitly capture pathology-specific morphological patterns that are critical for disease characterization. To address this limitation, we propose Tiny Vision Transformer with Pathology-Aware Prototype Distillation (TVT-PAPD). This self-supervised pathology representation learning framework integrates a Tiny Vision Transformer (TVT) with a novel Pathology-Aware Prototype Distillation (PAPD) module. PAPD employs a learnable pathology prototype bank to discover and preserve representative tissue morphology patterns, encouraging semantically similar pathological regions to learn consistent and discriminative representations. The proposed framework enhances pathology-aware feature learning while maintaining computational efficiency with 90M parameters. Experiments on the Cancer Genome Atlas (TCGA) low-grade glioma (LGG)/glioblastoma (GBM) dataset and the Indian Pathology Brain (IPD-Brain) dataset demonstrate that TVT-PAPD achieves weighted F1-scores of 93.02\% and 90.23\%, respectively, for LGG-GBM classification, while exhibiting strong cross-cohort generalization across independent glioma datasets.

Furthermore, the learned prototypes capture meaningful patterns in tissue morphology, improving the interpretability of the representation. These results highlight the effectiveness of pathology-aware prototype distillation for efficient, robust, and interpretable WSI analysis in computational pathology.
\end{abstract}


\section{Introduction} \label{sec:introduction}
Digital pathology has transformed computational pathology by enabling the digitization of histopathological slides into high-resolution WSIs~\cite{litjens2017survey}. These gigapixel-scale images contain rich morphological information for automated diagnosis, prognosis, and disease screening~\cite{bejnordi2017camelyon,hashimoto2020ai}. However, their massive size and tissue heterogeneity make efficient WSI analysis challenging~\cite{campanella2019clinical}. Furthermore, staining variability and limited expert annotations further complicate robust model development~\cite{jaume2021explainable,pinckaers2021detection}. Consequently, WSIs are typically processed as collections of image patches~\cite{campanella2019clinical}. Transformer-based architectures have substantially improved WSI analysis by modeling long-range contextual dependencies beyond the limited receptive fields of convolutional neural networks (CNNs). Recent approaches, including Vision Transformer (ViT)~\cite{dosovitskiy2021vit}, Clustering-constrained Attention Multiple Instance Learning (CLAM)~\cite{lu2021clam}, Transformer-based Multiple Instance Learning (TransMIL)~\cite{shao2021transmil}, and Hierarchical Image Pyramid Transformer (HIPT)~\cite{chen2022hipt}, have achieved strong performance in pathology image analysis, while pathology foundation models such as GigaPath~\cite{chen2023gigapath} and Virchow~\cite{lu2024virchow} have demonstrated the benefits of large-scale pretraining. Nevertheless, these models often require considerable computational resources.

SSL has emerged as an effective paradigm for learning transferable representations from unlabeled WSIs. Methods such as Simple Framework for Contrastive Learning of Visual Representations (SimCLR)~\cite{chen2020simclr}, Momentum Contrast (MoCo)~\cite{he2020moco}, Self-Distillation with No Labels (DINO)~\cite{caron2021dino}, and Self-Path~\cite{koohbanani2021selfpath} have demonstrated promising performance. However, most existing SSL methods primarily learn generic visual representations and do not explicitly preserve pathology-specific morphological patterns, limiting their ability to capture diagnostically meaningful tissue structures. To address this limitation, we propose \textbf{TVT-PAPD}, a self-supervised pathology representation learning framework that integrates a lightweight Tiny Vision Transformer (TVT) with a novel Pathology-Aware Prototype Distillation (PAPD) module. PAPD employs a learnable pathology prototype bank to model representative tissue morphology patterns and aligns teacher-student prototype distributions during self-supervised training. This enables morphology-aware representation learning while maintaining computational efficiency, resulting in more discriminative and interpretable pathology representations.

The main contributions of this work are as follows:
\begin{itemize}
\item \textbf{Pathology-Aware Prototype Distillation (PAPD):}
We propose a pathology-aware prototype distillation mechanism that integrates learnable pathology prototypes into self-supervised learning. By aligning teacher--student prototype distributions, PAPD preserves recurring tissue morphology patterns during pretraining without requiring pathology annotations.

\item \textbf{TVT-PAPD Framework:}
We develop a lightweight self-supervised WSI representation learning framework by integrating the Tiny Vision Transformer (TVT) with PAPD, enabling efficient and discriminative pathology representation learning with strong cross-cohort generalization.
\end{itemize}

\section{Methodology}\label{sec:methodology}
The proposed TVT-PAPD framework learns pathology-aware representations from gigapixel WSIs using SSL. Each WSI is partitioned into tissue patches, which are encoded by the Tiny Vision Transformer (TVT). The resulting representations are refined through the proposed Pathology-Aware Prototype Distillation (PAPD) module to preserve pathology-specific morphological patterns.

\subsection{Patch Extraction and Patch Embedding}
\begin{algorithm}[t]
\caption{Tissue Patch Selection for TVT-PAPD}
\label{alg:patch_extraction}
\begin{algorithmic}[1]

\REQUIRE Whole slide image $I$, patch size $P$, background threshold $\tau$
\ENSURE Tissue patch set $\mathcal{P}$

\STATE Initialize retained patch set $\mathcal{P}\leftarrow\emptyset$

\FOR{each grid location $(x,y)$ in $I$ with stride $P$}

    \STATE $t \leftarrow \mathrm{CropPatch}(I,x,y,P)$

    \STATE $\rho \leftarrow \mathrm{BackgroundRatio}(t)$

    \IF{$\rho < \tau$}
        \STATE $\mathcal{P}\leftarrow\mathcal{P}\cup\{t\}$
    \ENDIF

\ENDFOR

\RETURN $\mathcal{P}$

\end{algorithmic}
\end{algorithm}

Due to their gigapixel resolution, WSIs are partitioned into non-overlapping image patches for efficient processing. This representation preserves local tissue morphology while enabling scalable SSL.
Let $I \in \mathbb{R}^{H \times W \times C}$ denote a WSI, where $H$, $W$, and $C$ represent the image height, width, and number of channels, respectively. The image is partitioned into patches of size $P \times P$, yielding
\begin{equation}
N = \left\lfloor \frac{H}{P} \right\rfloor \times \left\lfloor \frac{W}{P} \right\rfloor.
\label{eq:num_patches}
\end{equation}

Each patch is defined as
\begin{equation}
p_i = I[x_i:x_i+P,\; y_i:y_i+P],
\label{eq:patch_def}
\end{equation}
where $(x_i,y_i)$ denotes the spatial location of the $i^{th}$ patch.

\begin{proposition}
The partitioning of a whole slide image into non-overlapping patches preserves its spatial structure up to patch-level resolution.
\end{proposition}

\paragraph{Paragraph:}
The image domain is partitioned into disjoint regions
$\{\Omega_i\}_{i=1}^{N}$ satisfying
$
\bigcup_{i=1}^{N}\Omega_i
=
\{1,\ldots,H\}\times\{1,\ldots,W\}
$
and
$
\Omega_i\cap\Omega_j=\emptyset
$
for $i\neq j$. Since each pixel belongs to exactly one patch, the spatial organization of the original WSI is preserved up to the chosen patch resolution.

To eliminate non-informative regions, patches with excessive background are discarded. A patch is retained only if:
\begin{equation}
\frac{\text{BackgroundPixels}(p_i)}{P^2}<\tau,
\label{eq:filter}
\end{equation}
where $\tau$ is a predefined threshold.

\begin{proposition}
The background filtering criterion improves the expected signal-to-noise ratio of the retained patch set.
\end{proposition}

\paragraph{Proof}
Background-dominant patches contain limited tissue information and therefore exhibit lower signal-to-noise ratios. Removing such patches suppresses background noise while preserving informative tissue regions, thereby increasing the expected signal-to-noise ratio of the retained patch set.

The retained patch set is defined as:
\begin{equation}
\mathcal{P}=\{p_1,p_2,\ldots,p_M\}, \quad M\le N,
\label{eq:patch_set}
\end{equation}
where $M$ denotes the number of retained patches. The complete extraction procedure is summarized in Algorithm~\ref{alg:patch_extraction}.

\subsubsection{Patch Embedding}
An embedding function $E(\cdot)$ is used to map each extracted patch $p_i$ to a $D$-dimensional embedding:
\begin{equation}
z_i = E(p_i).
\label{eq:embedding}
\end{equation}

To retain spatial information, positional encoding $e_i$ is added to the patch embedding:
\begin{equation}
\tilde{z}_i = z_i + e_i.
\label{eq:positional}
\end{equation}

The resulting token sequence is provided as input to the Tiny Vision Transformer (TVT), which learns contextual representations of tissue morphology through self-attention mechanisms. These representations subsequently serve as the basis for pathology-aware prototype learning within the proposed PAPD framework.

\subsection{Tiny Vision Transformer (TVT) Backbone}
\begin{figure}[!t]
\centering
\includegraphics[width=0.5\columnwidth]{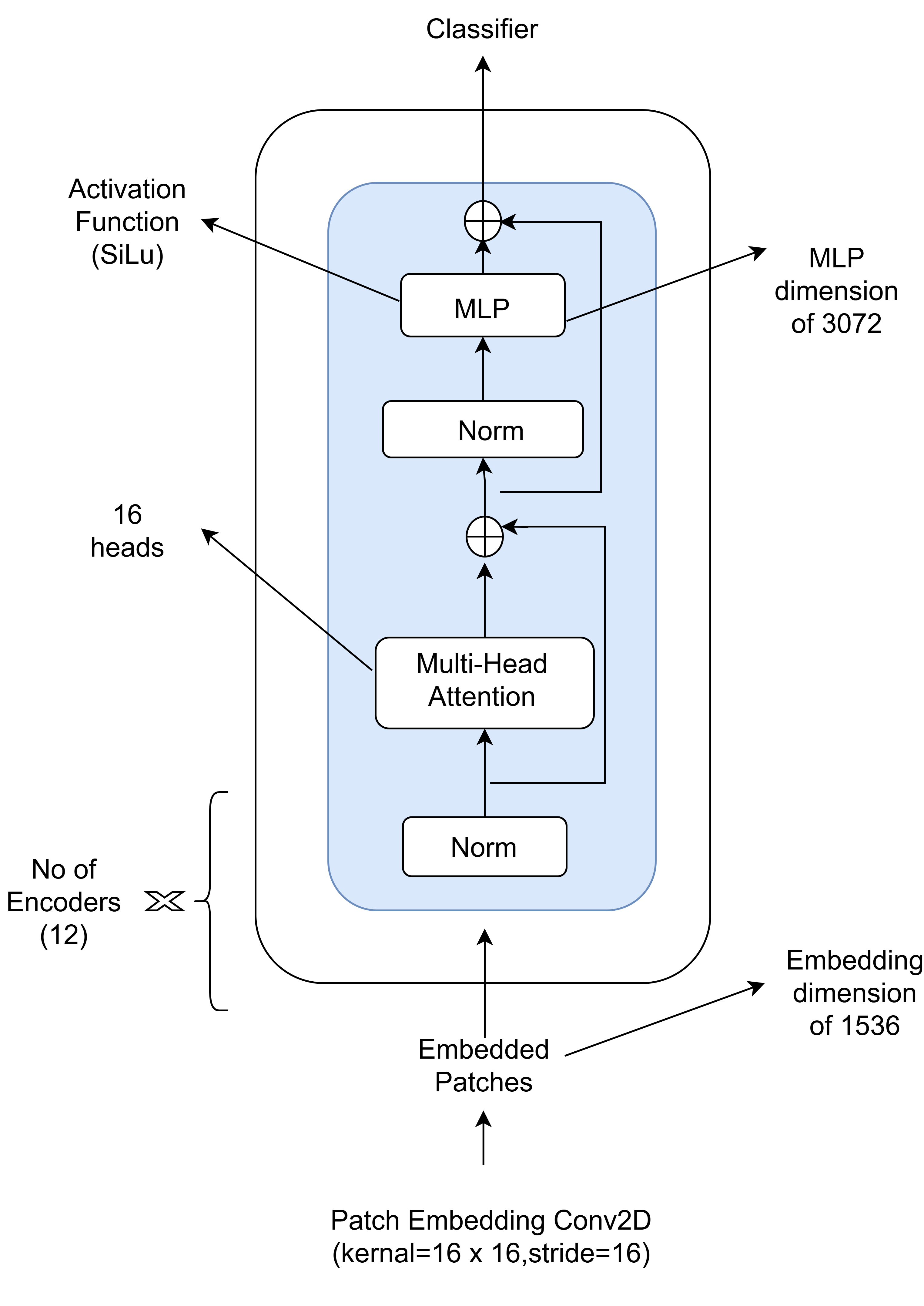} 
\caption{Architecture of the proposed Vision Transformer backbone.}
\label{fig:arch}
\end{figure}

The proposed framework employs a customized Tiny Vision Transformer (TVT) backbone, illustrated in Fig.~\ref{fig:arch}. The network consists of 12 transformer encoder layers with 16-head multi-head self-attention. Input patches are embedded using a Conv2D patch embedding layer (kernel size 16, stride 16) and projected into a 1536-dimensional feature space. Each encoder block comprises layer normalization, multi-head self-attention, and an MLP with a hidden dimension of 3072 and SiLU activation, enabling efficient modeling of both local tissue morphology and global contextual dependencies.

Each token representation updated using the self-attention mechanism using attention weights $\alpha_{ij}$ is:
\begin{equation}
  z_i' = \sum_{j=1}^{M} \alpha_{ij} z_j, \text{ where }
  \alpha_{ij} =
\mathrm{softmax}
\left(
\frac{q_i^{T}k_j}{\sqrt{d}}
\right),
\label{eq:self_attention}
\end{equation}
where $q_i$, $k_j$, and $z_j$ denote the query, key, and value representations, respectively.

\begin{proposition}
The self-attention mechanism enables global interactions among all input tokens within a single layer.
\end{proposition}

\paragraph{Proof}
Each output token $z_i'$ is computed as a weighted combination of all input tokens, where the attention weights satisfy $\alpha_{ij}>0$ and $\sum_j\alpha_{ij}=1$. Consequently, every token aggregates information from the entire input sequence within a single layer, providing a global receptive field and enabling the modeling of long-range spatial dependencies.

\subsection{Self-Supervised Learning with DINO}
The proposed framework adopts the DINO self-supervised learning paradigm based on a student--teacher architecture. Following the TVT backbone, a standard DINO projection head consisting of a two-layer MLP followed by a linear projection maps patch representations into a high-dimensional embedding space for self-supervised learning. The projection head configuration is summarized in Table~\ref{tab:dino_head}.

Multiple augmented views of each image patch are generated, and the student network is optimized to match the teacher predictions, enabling invariant representation learning from unlabeled histopathology images. The DINO objective is:
\begin{equation}
\mathcal{L}_{\mathrm{DINO}}
=
-\sum_i p_t(x_i)\log p_s(x_i),
\label{eq:dino_loss}
\end{equation}
where $p_t(x_i)$ and $p_s(x_i)$ denote the teacher and student output distributions. The teacher parameters are updated using an exponential moving average (EMA):
\begin{equation}
\theta_t \leftarrow m\theta_t + (1-m)\theta_s,
\label{eq:ema}
\end{equation}
where $\theta_t$ and $\theta_s$ denote the teacher and student parameters, and $m$ is the momentum coefficient.

\begin{table}[t]
\caption{Architecture configuration of the DINO projection head.}
\label{tab:dino_head}
\centering
\small
\renewcommand{\arraystretch}{1.3}
\setlength{\tabcolsep}{6pt}
\begin{tabular}{cc}
\hline
\textbf{Component} & \textbf{Dimension} \\
\hline
Input Feature & 1536 \\
FC Layer 1 & 2048 \\
FC Layer 2 & 256 \\
Output Projection & 8192 \\
Activation Function & SiLU \\
Projection Type & Linear \\
\hline
\end{tabular}
\end{table}

\subsection{Pathology-Aware Prototype Distillation (PAPD)}\label{sec:papd}
\begin{figure*}[t]
\centering
\includegraphics[width=\textwidth]{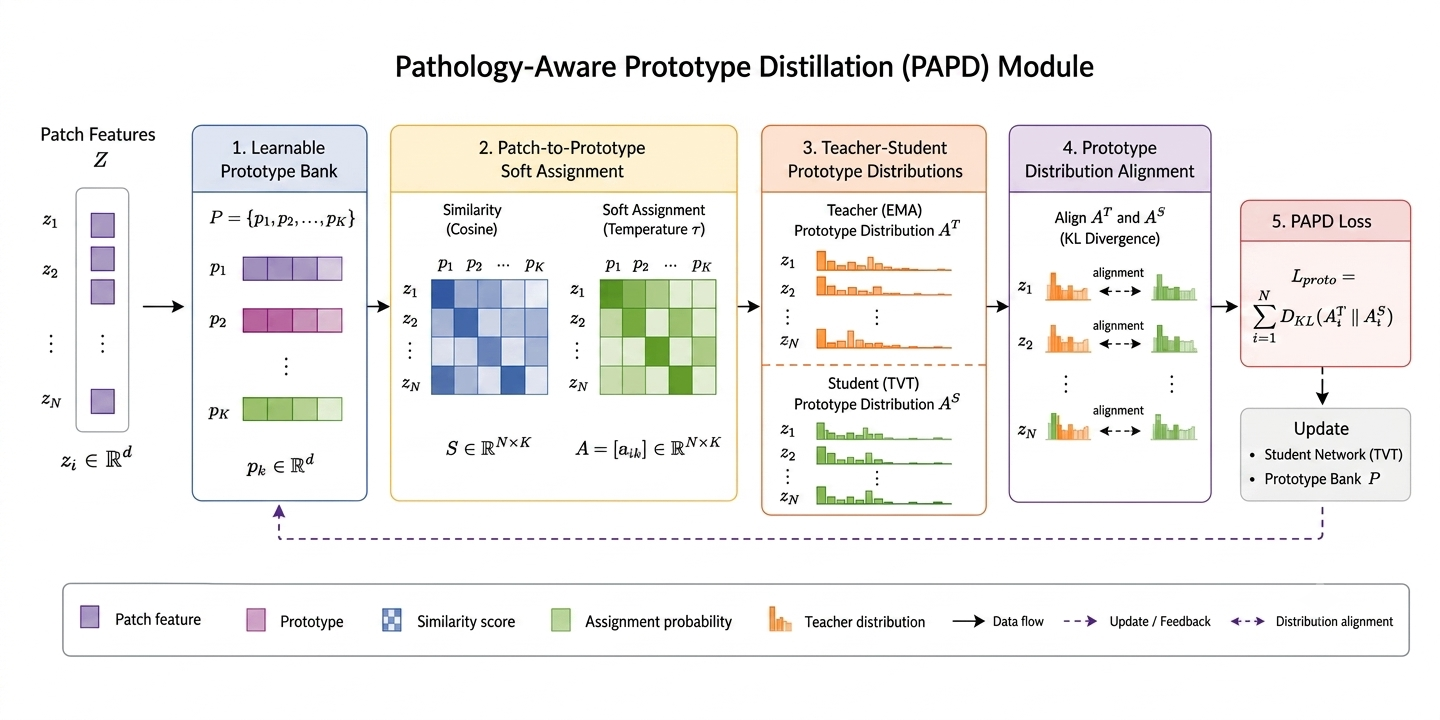}
\caption{Architecture of the proposed Pathology-Aware Prototype Distillation (PAPD) module. Patch features are associated with a learnable pathology prototype bank to obtain soft prototype assignments. Teacher and student prototype distributions are aligned through prototype distillation, enabling morphology-aware representation learning by preserving representative tissue patterns during self-supervised training.}
\label{fig:papd}
\end{figure*}

Although DINO learns robust visual representations, it does not explicitly preserve pathology-specific morphology. To address this limitation, we propose Pathology-Aware Prototype Distillation (PAPD), which integrates a learnable pathology prototype bank into the teacher--student self-supervised framework. As illustrated in Fig.~\ref{fig:papd}, patch representations are softly assigned to pathology prototypes, and the teacher and student prototype distributions are aligned to encourage morphology-aware representation learning.

\subsubsection{Motivation for Pathology-Aware Prototype Learning}
Histopathology images contain recurring tissue morphology, including cellular organization and tissue architecture, that is critical for diagnosis. However, existing self-supervised learning methods primarily optimize generic visual representations without explicitly preserving these pathology-specific patterns. To address this limitation, PAPD introduces a learnable pathology prototype bank that models representative tissue morphology. By aligning teacher--student prototype distributions during self-supervised training, the proposed framework encourages morphologically similar tissue regions to learn consistent and discriminative representations.

\subsubsection{Learnable Pathology Prototype Bank}
Let
\begin{equation}
Z={z_1,z_2,\ldots,z_N},
\end{equation}
denote the patch representations extracted by the TVT backbone, where $z_i \in \mathbb{R}^{d}$. A learnable pathology prototype bank $P$, to capture representative tissue morphology patterns, is:
\begin{equation}
P={p_1,p_2,\ldots,p_K},
\end{equation}
where $p_k \in \mathbb{R}^{d}$ represents the $k^{th}$ pathology prototype and $K$ denotes the number of prototypes. The prototype bank acts as a morphology-aware memory that captures recurring pathological structures across whole slide images.

Unlike fixed clustering centroids, the prototype bank is jointly optimized with the network parameters during training. Through repeated teacher-student prototype alignment, the prototypes gradually adapt to the evolving feature distribution and capture representative tissue morphology patterns that frequently occur across the training dataset.

\subsubsection{Patch-to-Prototype Association}
For each patch representation $z_i$, cosine similarity is computed with all prototypes:
\begin{equation}
s_{ik}
=
\frac{z_i^{T}p_k}
{\|z_i\|\|p_k\|}.
\end{equation}

The similarity scores are transformed into soft prototype assignments using a temperature-scaled softmax:
\begin{equation}
a_{ik}
=
\frac{\exp(s_{ik}/\tau)}
{\sum_{j=1}^{K}\exp(s_{ij}/\tau)},
\end{equation}
where $\tau$ is the temperature parameter. The assignment probability $a_{ik}$ represents the association between patch $i$ and prototype $k$, enabling each patch to be represented by multiple morphology patterns through soft assignments.

\subsubsection{Pathology-Aware Representation Learning}
Using the prototype assignment probabilities and prototype embeddings, a pathology-aware representation is given as a weighted sum:
\begin{equation}
r_i
=
\sum_{k=1}^{K}
a_{ik}p_k.
\end{equation}
The resulting representation enriches the transformer features with morphology-aware information derived from the learned pathology prototypes.

\subsubsection{Prototype Representation Learning}
Let $A^{T}=\{a_{ik}^{T}\}$ and $A^{S}=\{a_{ik}^{S}\}$ denote the teacher and student prototype assignment distributions, respectively. To preserve morphology-aware structures during self-supervised learning, the teacher and student distributions are aligned using a prototype distillation objective:
\begin{equation}
L_{proto}
=
\sum_{i=1}^{N}
D_{KL}
\left(
A_i^{T}
\Vert
A_i^{S}
\right)
=
\sum_{i=1}^{N}
\sum_{k=1}^{K}
a_{ik}^{T}
\log
\frac{a_{ik}^{T}}
{a_{ik}^{S}}.
\end{equation}

Minimizing $L_{proto}$ encourages the student network to preserve the pathology prototype structure learned by the teacher, resulting in more discriminative pathology representations.

\subsection{Joint Optimization Objective}
The proposed TVT-PAPD jointly optimizes the DINO self-supervised objective and the proposed Pathology-Aware Prototype Distillation (PAPD) objective. The DINO loss promotes feature-level consistency between teacher and student representations, while the PAPD loss preserves pathology-specific morphology through prototype distribution alignment.

The overall training objective is defined as
\begin{equation}
L_{\text{total}}
=
L_{\text{DINO}}
+
\lambda L_{\text{proto}},
\label{eq:total_loss}
\end{equation}
where $L_{\text{DINO}}$ is the DINO self-supervised loss, $L_{\text{proto}}$ is the proposed prototype distillation loss, and $\lambda$ balances the contribution of prototype supervision. Joint optimization enables the model to learn discriminative pathology representations by capturing both generic visual semantics and pathology-specific morphological structures.

\subsection{Framework Overview and Training Algorithm}
\begin{figure}[!t]
\centering
\includegraphics[width=0.75\columnwidth]{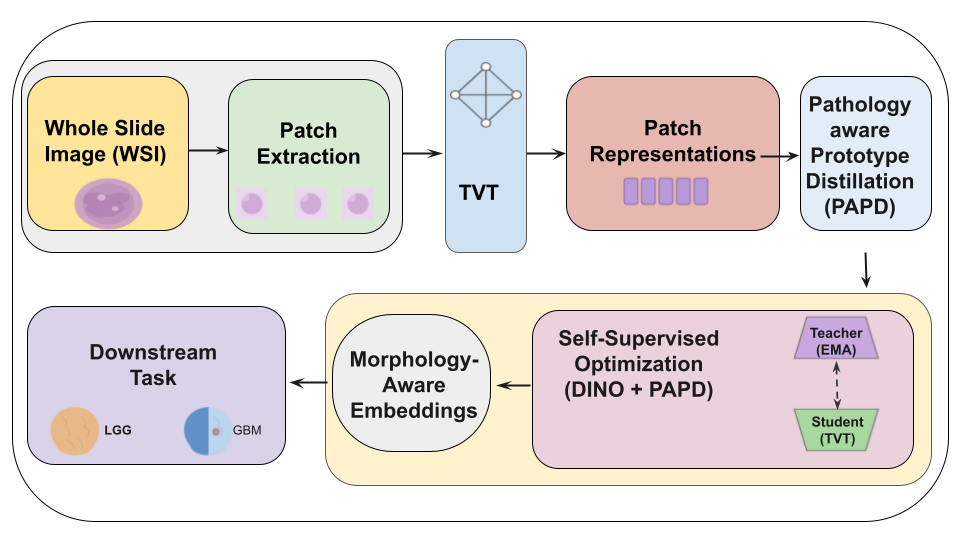}
\caption{Training pipeline of the proposed TVT-PAPD model. Tissue patches extracted from whole slide images are encoded by the Tiny Vision Transformer (TVT) and jointly optimized using the DINO self-supervised objective and the proposed Pathology-Aware Prototype Distillation (PAPD) objective to learn morphology-aware pathology representations.}
\label{fig:pipeline}
\end{figure}

Figure~\ref{fig:pipeline} illustrates the overall training pipeline of the proposed TVT-PAPD framework, while Algorithm~\ref{alg:tvt_papd} summarizes the corresponding optimization procedure. Given a mini-batch of tissue patches, augmented views are processed by the student and teacher networks sharing the same TVT backbone. The student network is optimized using the joint DINO and PAPD objectives, whereas the teacher parameters are updated through an exponential moving average (EMA) of the student parameters. Prototype distributions from both networks are aligned via the PAPD loss, enabling morphology-aware self-supervised representation learning without manual annotations.

\begin{algorithm}[t]
\caption{TVT-PAPD Self-Supervised Training}
\label{alg:tvt_papd}
\begin{algorithmic}[1]

\REQUIRE
$\mathcal{P}$,
$P$,
$\theta_s$,
$\theta_t$,
$\eta$,
$m$

\FOR{each mini-batch $\mathcal{B}\subset\mathcal{P}$}

\STATE $(v_s,v_t)\leftarrow\mathcal{T}(\mathcal{B})$

\STATE $Z_s\leftarrow f_{\theta_s}(v_s)$

\STATE $Z_t\leftarrow f_{\theta_t}(v_t)$

\STATE $L_{\mathrm{DINO}}
\leftarrow
-\sum_i p_t(x_i)\log p_s(x_i)$

\STATE $A^S\leftarrow\mathrm{Softmax}(Z_sP^\top)$

\STATE $A^T\leftarrow\mathrm{Softmax}(Z_tP^\top)$

\STATE $L_{\mathrm{proto}}
\leftarrow
\sum_i
D_{\mathrm{KL}}
(A_i^T\Vert A_i^S)$

\STATE $L_{\mathrm{total}}
\leftarrow
L_{\mathrm{DINO}}
+
\lambda L_{\mathrm{proto}}$

\STATE $\theta_s
\leftarrow
\theta_s
-\eta\nabla_{\theta_s}
L_{\mathrm{total}}$

\STATE $\theta_t
\leftarrow
m\theta_t
+(1-m)\theta_s$

\ENDFOR

\RETURN $\theta_s$

\end{algorithmic}
\end{algorithm}

\section{Experimental Setup}\label{sec:expt}
This section describes the dataset, preprocessing pipeline, training configuration, implementation details, and evaluation protocol used to validate the proposed TVT framework. The experiments are designed to assess the effectiveness of the proposed self-supervised learning approach for representation learning on large-scale WSIs.

\subsection{Dataset}
The proposed TVT-PAPD framework is evaluated on glioma WSIs from The Cancer Genome Atlas (TCGA)~\cite{tcga}, including the TCGA-LGG and TCGA-GBM cohorts, obtained through the Genomic Data Commons (GDC) Data Portal~\cite{heath2021nci,gdc}. To evaluate cross-dataset generalization, experiments are further conducted on the publicly available IPD-Brain dataset~\cite{chauhan2024ipdbrain}. For consistency, astrocytoma and oligodendroglioma cases are grouped as LGG, while glioblastoma cases are assigned to the GBM category. The complete dataset statistics are summarized in Table~\ref{tab:dataset_stats}.

\begin{table}[t]
\caption{Summary of datasets used in this study.}
\label{tab:dataset_stats}
\centering
\small
\renewcommand{\arraystretch}{1.3}
\setlength{\tabcolsep}{6pt}
\begin{tabular}{cc}
\hline
\textbf{Property} & \textbf{Value} \\
\hline
Training Dataset & TCGA-LGG, TCGA-GBM \\
Training Patches & 1,977,173 \\
Validation Patches & 40,840 \\
\hline
TCGA Test Dataset & TCGA-LGG, TCGA-GBM \\
TCGA Test Patches & 368,000 \\
\hline
IPD-Brain Test Dataset & IPD-Brain \\
IPD-Brain Test Patches & 275,000 \\
\hline
Patch Resolution & $224 \times 224$ \\
Image Channels & RGB \\
Patch Extraction & Sliding Window \\
\hline
Training Paradigm & Self-Supervised Learning \\
Primary Task & LGG vs GBM Classification \\
IPD-Brain Test Task & LGG vs GBM Classification  \\
\hline
\end{tabular}
\end{table}

\subsection{Hardware Configuration}
All experiments are conducted on a high-performance computing system with GPU acceleration support. To ensure consistent and reproducible results, all experiments are performed using a single GPU.

The model is implemented using the PyTorch deep learning framework and trained with mixed precision to improve computational efficiency and reduce memory consumption. Gradient accumulation is employed to effectively handle large-scale training under memory constraints. Detailed hardware and software configurations are summarized in Table~\ref{tab:hardware_details}.

\begin{table}[t]
\caption{Hardware and software configuration used in the experiments.}
\label{tab:hardware_details}
\small
\renewcommand{\arraystretch}{1.1}
\setlength{\tabcolsep}{6pt}
\centering
\begin{tabular}{cc}
\hline
\textbf{Component} & \textbf{Specification} \\
\hline
CPU & Dual AMD EPYC 9534 (64 cores \\
& per socket, 256 threads total) \\
CPU Architecture & x86\_64 \\
\hline
GPU & AMD Instinct Accelerators \\
& (ROCm-enabled) \\
GPU Count & 8 (1 GPU used for training) \\
GPU Memory & 192 GB HBM (per GPU) \\
GPU Framework & ROCm \\
\hline
Deep Learning Framework & PyTorch \\
Training Precision & Mixed Precision \\
Parallelization Strategy & Gradient Accumulation \\
\hline
\end{tabular}
\end{table}

\subsection{Training Details}
The proposed TVT-PAPD framework is implemented in PyTorch and trained using GPU acceleration with mixed-precision training to improve computational efficiency. The Tiny Vision Transformer (TVT) backbone is optimized within the DINO teacher--student self-supervised framework, while the proposed Pathology-Aware Prototype Distillation (PAPD) module is jointly trained to preserve pathology-specific morphological patterns. The teacher network is updated using an exponential moving average (EMA) of the student parameters, and the overall model is optimized using the joint objective defined in Eq.~(\ref{eq:total_loss}). The complete training hyperparameter configuration is summarized in Table~\ref{tab:training_config}.

\begin{table}[t]
\caption{Training configuration used for model optimization.}
\label{tab:training_config}
\centering
\small
\renewcommand{\arraystretch}{1.3}
\setlength{\tabcolsep}{6pt}
\begin{tabular}{lc|lc}
\hline
\textbf{Parameter} & \textbf{Value}
& \textbf{Parameter} & \textbf{Value}\\
\hline
Optimizer & AdamW 
& Steps per Epoch & 30,893 \\

Batch Size & 64 
& Gradient Accumulation & 4 \\

Weight Decay & $10^{-4}$
& Mixed Precision Training & Enabled \\

Epochs & 50 
& Teacher Momentum ($m$) & 0.996 \\

Learning Rate & $10^{-4}$ 
& Num. of Prototypes ($K$) & 64 \\

Global Crop & 224
& Temperature ($\tau$) & 0.1 \\

Size & 
& PAPD Loss Weight ($\lambda$) & 0.5 \\
\hline
\end{tabular}
\end{table}

\section{Results and Discussion}\label{sec:results}
This section presents the experimental evaluation of the proposed TVT-PAPD framework on the TCGA glioma and IPD-Brain datasets. The learned representations are assessed through downstream LGG-GBM classification, cross-cohort evaluation, ablation studies, representation quality analysis, and prototype interpretability experiments to evaluate the effectiveness of the proposed pathology-aware prototype distillation mechanism.

\subsection{Classification Performance}
Table~\ref{tab:classfn_results} summarizes the classification performance of the proposed TVT-PAPD framework on the TCGA glioma and IPD-Brain datasets. The proposed method achieves high precision, recall, and F1-score across both datasets, demonstrating robust classification performance and strong cross-cohort generalization for LGG-GBM classification.

\begin{table}[t]
  \centering
  \caption{Classification performance of the proposed TVT-PAPD framework}
  \label{tab:classfn_results}
  \small
  \renewcommand{\arraystretch}{1.3}
  \setlength{\tabcolsep}{6pt}

  \begin{tabular}{lcccc}
    \hline
    \bf Class & \bf Precision & \bf Recall & \bf F1 & \bf Support \\\hline
    \multicolumn{5}{c}{\textit{TCGA Glioma Test Dataset} (\textbf{Overall Accuracy 0.932})} \\\hline
    LGG & 0.934 & 0.920 & 0.927 & 181,569 \\
    GBM & 0.927 & 0.940 & 0.933 & 186,411 \\
    Macro $\mu$ & 0.931 & 0.930 & 0.930 & 368,000 \\
    Weighted $\mu$ & 0.930 & 0.932 & 0.930 & 368,000 \\ \hline
    \multicolumn{5}{c}{\textit{IPD Test Data} (\textbf{Overall Accuracy 0.917})}\\\hline
    LGG & 0.897 & 0.884 & 0.891 & 125,000 \\
    GBM & 0.905 & 0.919 & 0.912 & 150,000 \\
    Macro $\mu$ & 0.901 & 0.902 & 0.901 & 275,000 \\
    Weighted $\mu$ & 0.902 & 0.917 & 0.902 & 275,000 \\\hline
  \end{tabular}
\end{table}

\subsection{Comparison with State-of-the-Art Methods}
We compare the proposed TVT-PAPD framework with representative CNN-based, transformer-based, and pathology foundation models to evaluate its effectiveness on the TCGA-LGG/GBM and IPD-Brain datasets. As shown in Table~\ref{tab:sota_comparison}, the proposed method achieves the best overall performance, demonstrating the effectiveness of pathology-aware prototype distillation for histopathology representation learning.

\begin{table*}[t]
\caption{Comparison with state-of-the-art whole slide image classification methods on the TCGA-LGG/GBM and IPD-Brain datasets.}
\label{tab:sota_comparison}
\centering
\scriptsize
\renewcommand{\arraystretch}{1.3}
\setlength{\tabcolsep}{6pt}

\begin{tabular}{l|c|ccc|ccc}
\hline
\multirow{2}{*}{\textbf{Method}} &
\multirow{2}{*}{\textbf{Backbone}} &
\multicolumn{3}{c|}{\textbf{TCGA-LGG/GBM}} &
\multicolumn{3}{c}{\textbf{IPD-Brain}}\\
\cline{3-8}
&
&
\textbf{Acc.} &
\textbf{F1} &
\textbf{AUC} &
\textbf{Acc.} &
\textbf{F1} &
\textbf{AUC}\\
\hline

CLAM~\cite{lu2021clam}
&
ResNet-50
&
88.41 $\pm$ 0.48
&
88.12 $\pm$ 0.46
&
0.93 $\pm$ 0.02
&
86.72 $\pm$ 0.51
&
86.45 $\pm$ 0.49
&
0.91 $\pm$ 0.02
\\

TransMIL~\cite{shao2021transmil}
&
Vision Transformer
&
89.36 $\pm$ 0.42
&
89.05 $\pm$ 0.40
&
0.94 $\pm$ 0.02
&
87.51 $\pm$ 0.47
&
87.24 $\pm$ 0.45
&
0.92 $\pm$ 0.02
\\

HIPT~\cite{chen2022hipt}
&
Hierarchical ViT
&
91.18 $\pm$ 0.36
&
90.94 $\pm$ 0.34
&
0.96 $\pm$ 0.01
&
89.43 $\pm$ 0.39
&
89.12 $\pm$ 0.37
&
0.95 $\pm$ 0.01
\\

CTransPath~\cite{wang2022ctranspath}
&
Swin Transformer
&
91.67 $\pm$ 0.34
&
91.42 $\pm$ 0.32
&
0.96 $\pm$ 0.01
&
89.96 $\pm$ 0.36
&
89.71 $\pm$ 0.35
&
0.95 $\pm$ 0.01
\\

RetCCL~\cite{zhou2022retccl}
&
ResNet-50
&
91.95 $\pm$ 0.33
&
91.73 $\pm$ 0.31
&
0.96 $\pm$ 0.01
&
90.38 $\pm$ 0.34
&
90.02 $\pm$ 0.32
&
0.95 $\pm$ 0.01
\\

GigaPath~\cite{chen2023gigapath}
&
Vision Transformer
&
92.68 $\pm$ 0.32
&
92.41 $\pm$ 0.30
&
0.97 $\pm$ 0.01
&
91.02 $\pm$ 0.33
&
90.76 $\pm$ 0.31
&
0.96 $\pm$ 0.01
\\

\textbf{Proposed}
&
\textbf{Tiny Vision Transformer}
&
\textbf{93.25 $\pm$ 0.31}
&
\textbf{93.02 $\pm$ 0.28}
&
\textbf{0.97 $\pm$ 0.01}
&
\textbf{91.71 $\pm$ 0.42}
&
\textbf{90.23 $\pm$ 0.37}
&
\textbf{0.96 $\pm$ 0.01}
\\
\hline

\end{tabular}
\end{table*}

\subsection{Comparison with Self-Supervised Learning Methods}
To evaluate the effectiveness of the proposed Pathology-Aware Prototype Distillation (PAPD) strategy, we compare it with several representative self-supervised learning approaches using the same TVT backbone. The comparison includes contrastive learning methods (SimCLR and MoCo v3), self-distillation approaches (BYOL, DINO, and DINOv2), and masked image modeling techniques (MAE and iBOT). All methods were evaluated using the same training and testing protocol on both the TCGA-LGG/GBM and IPD-Brain datasets. Each experiment was repeated three times with different random seeds, and the mean performance together with the standard deviation is reported. The proposed TVT-PAPD framework consistently outperformed competing SSL methods with low performance variance.

Table~\ref{tab:comparison} summarizes the results. The proposed TVT-PAPD framework consistently outperforms existing SSL methods across both datasets, demonstrating the effectiveness of pathology-aware prototype learning for robust and transferable histopathological representation learning.

\begin{table*}[t]
\caption{Comparison of different self-supervised learning strategies using the same TVT backbone and representative prototype-based learning approaches on the TCGA glioma and IPD-Brain datasets.}
\label{tab:comparison}
\centering
\small
\renewcommand{\arraystretch}{1.1}
\setlength{\tabcolsep}{8pt}

\begin{tabular}{l|ccc|ccc}
  \hline
& \multicolumn{3}{c|}{\textbf{TCGA-LGG/GBM}}
& \multicolumn{3}{c}{\textbf{IPD-Brain}} \\
\cline{2-7}
\bf Method 
& \textbf{Acc.} & \textbf{F1} & \textbf{AUC}
& \textbf{Acc.} & \textbf{F1} & \textbf{AUC} \\
\hline
\multicolumn{7}{c}{\bf Self-Supervised Learning with TVT Backbone} \\\hline
TVT + SimCLR
& 86.73 $\pm$ 0.62
& 86.51 $\pm$ 0.58
& 0.91 $\pm$ 0.02
& 84.82 $\pm$ 0.71
& 84.56 $\pm$ 0.65
& 0.89 $\pm$ 0.02 \\

TVT + MoCo v3
& 87.84 $\pm$ 0.56
& 87.65 $\pm$ 0.53
& 0.92 $\pm$ 0.02
& 85.91 $\pm$ 0.68
& 85.74 $\pm$ 0.61
& 0.90 $\pm$ 0.02 \\

TVT + BYOL
& 88.42 $\pm$ 0.51
& 88.19 $\pm$ 0.47
& 0.93 $\pm$ 0.02
& 86.57 $\pm$ 0.64
& 86.31 $\pm$ 0.59
& 0.91 $\pm$ 0.02 \\

TVT + DINO
& 89.76 $\pm$ 0.48
& 89.58 $\pm$ 0.45
& 0.95 $\pm$ 0.01
& 87.88 $\pm$ 0.58
& 87.63 $\pm$ 0.54
& 0.93 $\pm$ 0.01 \\

TVT + DINOv2
& 90.48 $\pm$ 0.44
& 90.31 $\pm$ 0.41
& 0.96 $\pm$ 0.01
& 88.54 $\pm$ 0.53
& 88.26 $\pm$ 0.49
& 0.94 $\pm$ 0.01 \\

TVT + MAE
& 90.21 $\pm$ 0.46
& 90.03 $\pm$ 0.43
& 0.95 $\pm$ 0.01
& 88.17 $\pm$ 0.55
& 87.95 $\pm$ 0.51
& 0.93 $\pm$ 0.01 \\

TVT + iBOT
& 91.12 $\pm$ 0.39
& 90.96 $\pm$ 0.36
& 0.96 $\pm$ 0.01
& 89.42 $\pm$ 0.45
& 89.11 $\pm$ 0.41
& 0.95 $\pm$ 0.01 \\
\hline

\hline
\multicolumn{7}{c}{\bf Representative Prototype-based Learning}\\
\hline

DINO
& 89.76 $\pm$ 0.48
& 89.58 $\pm$ 0.45
& 0.95 $\pm$ 0.01
& 87.88 $\pm$ 0.58
& 87.63 $\pm$ 0.54
& 0.93 $\pm$ 0.01 \\

SwAV
& 90.34 $\pm$ 0.44
& 90.12 $\pm$ 0.41
& 0.95 $\pm$ 0.01
& 88.46 $\pm$ 0.52
& 88.18 $\pm$ 0.49
& 0.94 $\pm$ 0.01 \\

PANTHER
& 91.47 $\pm$ 0.39
& 91.18 $\pm$ 0.36
& 0.96 $\pm$ 0.01
& 89.52 $\pm$ 0.47
& 89.23 $\pm$ 0.44
& 0.95 $\pm$ 0.01 \\

ProtoMIL
& 92.14 $\pm$ 0.35
& 91.89 $\pm$ 0.33
& 0.96 $\pm$ 0.01
& 90.24 $\pm$ 0.43
& 89.96 $\pm$ 0.40
& 0.95 $\pm$ 0.01 \\
\hline

\textbf{TVT + DINO + PAPD (Ours)}
& \textbf{93.25 $\pm$ 0.31}
& \textbf{93.02 $\pm$ 0.28}
& \textbf{0.97 $\pm$ 0.01}
& \textbf{91.71 $\pm$ 0.42}
& \textbf{90.23 $\pm$ 0.37}
& \textbf{0.96 $\pm$ 0.01} \\
\hline

\end{tabular}
\end{table*}

\subsection{Comparison with Prototype-Based Learning Methods}
We compare the proposed TVT-PAPD framework with representative prototype-based learning methods, including DINO, SwAV, PANTHER, and ProtoMIL, using the same training and evaluation protocol on the TCGA-LGG/GBM and IPD-Brain datasets. As shown in Table~\ref{tab:comparison}, TVT-PAPD achieves the best performance across both datasets. Compared with DINO and SwAV, the proposed framework benefits from pathology-aware prototype learning, while outperforming PANTHER and ProtoMIL by integrating prototype distillation directly into the teacher--student self-supervised optimization process. These results demonstrate the effectiveness of the proposed PAPD module in learning discriminative pathology representations.

\subsection{Ablation Study}
We perform ablation studies to evaluate the contributions of PAPD, the number of pathology prototypes, and the PAPD loss weight.

\subsubsection{Effect of Pathology-Aware Prototype Distillation}
Table~\ref{tab:ablation_papd} presents the ablation results of the proposed PAPD module. Incorporating PAPD consistently improves performance over the TVT and TVT+DINO baselines, while the complete TVT+DINO+PAPD framework achieves the best results across both datasets.

\begin{table*}[t]
\caption{Ablation study of the proposed PAPD module on the TCGA glioma and IPD-Brain datasets.}
\label{tab:ablation_papd}
\centering
\small
\renewcommand{\arraystretch}{1.3}
\setlength{\tabcolsep}{6pt}
\begin{tabular}{ccc|ccc|ccc}
\hline
&&&
\multicolumn{3}{c|}{\textbf{TCGA-LGG/GBM}} &
\multicolumn{3}{c}{\textbf{IPD-Brain}} \\
\cline{4-9}
\bf TVT & \bf DINO & \bf PAPD
& \textbf{Acc.} & \textbf{F1} & \textbf{AUC}
& \textbf{Acc.} & \textbf{F1} & \textbf{AUC} \\
\hline
$\checkmark$ & $\times$ & $\times$
& 85.12 $\pm$ 0.68 & 84.91 $\pm$ 0.63 & 0.89 $\pm$ 0.02
& 83.94 $\pm$ 0.72 & 83.72 $\pm$ 0.67 & 0.88 $\pm$ 0.02 \\
$\checkmark$ & $\checkmark$ & $\times$ 
& 89.76 $\pm$ 0.48 & 89.58 $\pm$ 0.45 & 0.95 $\pm$ 0.01 
& 87.88 $\pm$ 0.58 & 87.63 $\pm$ 0.54 & 0.93 $\pm$ 0.01 \\
$\checkmark$ & $\times$ & $\checkmark$
& 90.88 $\pm$ 0.42 & 90.71 $\pm$ 0.39 & 0.96 $\pm$ 0.01 
& 89.21 $\pm$ 0.49 & 88.96 $\pm$ 0.45 & 0.95 $\pm$ 0.01 \\
\hline
$\checkmark$ & $\checkmark$ & $\checkmark$ &
\textbf{93.25 $\pm$ 0.31} &
\textbf{93.02 $\pm$ 0.28} &
\textbf{0.97 $\pm$ 0.01} &
\textbf{91.71 $\pm$ 0.42} &
\textbf{90.23 $\pm$ 0.37} &
\textbf{0.96 $\pm$ 0.01} \\
\hline
\end{tabular}
\end{table*}

\subsubsection{Effect of Number of Pathology Prototypes}
To analyze the influence of the pathology prototype bank size, we evaluate the proposed TVT-PAPD framework using different numbers of prototypes ($K$). As shown in Table~\ref{tab:parameter_experiments}, performance improves as the number of prototypes increases from 16 to 64, indicating that a larger prototype bank captures more diverse tissue morphology patterns. However, increasing $K$ to 128 provides limited additional benefit and slightly reduces performance. These results suggest that $K=64$ offers an effective balance between representation capacity and generalization.

\begin{table}[t]
  \caption{Experiments with number of pathology prototypes ($K$) and PAPD loss weight ($\lambda$).}
  \label{tab:parameter_experiments}
  \centering
  \small
  \begin{tabular}{c|cccc}
    \hline
    $K$ & 16 & 32 & 64 & 128 \\ 
    Acc. (\%)
    & 89.94 $\pm$ 0.52 & 90.71 $\pm$ 0.44
    & \bf 93.25 $\pm$ 0.31 & 91.11 $\pm$ 0.36 \\
    Wtd. F1 (\%)
    & 90.12 $\pm$ 0.48 & 91.03 $\pm$ 0.41
    & \bf 93.02 $\pm$ 0.28 & 91.82 $\pm$ 0.36 \\
    \hline
    $\lambda$ & 0.1 & 0.3 & 0.5 & 1.0 \\
    Acc. (\%)
    & 90.32 $\pm$ 0.51 & 90.81 $\pm$ 0.43
    & \bf 93.25 $\pm$ 0.31 & 90.73 $\pm$ 0.45 \\
    Wtd. F1 (\%)
    & 90.58 $\pm$ 0.47 & 91.37 $\pm$ 0.40
    & \bf 93.02 $\pm$ 0.28 & 91.18 $\pm$ 0.42 \\
    \hline
  \end{tabular}
\end{table}

\begin{figure*}[t]
\centering
\includegraphics[width=\columnwidth]{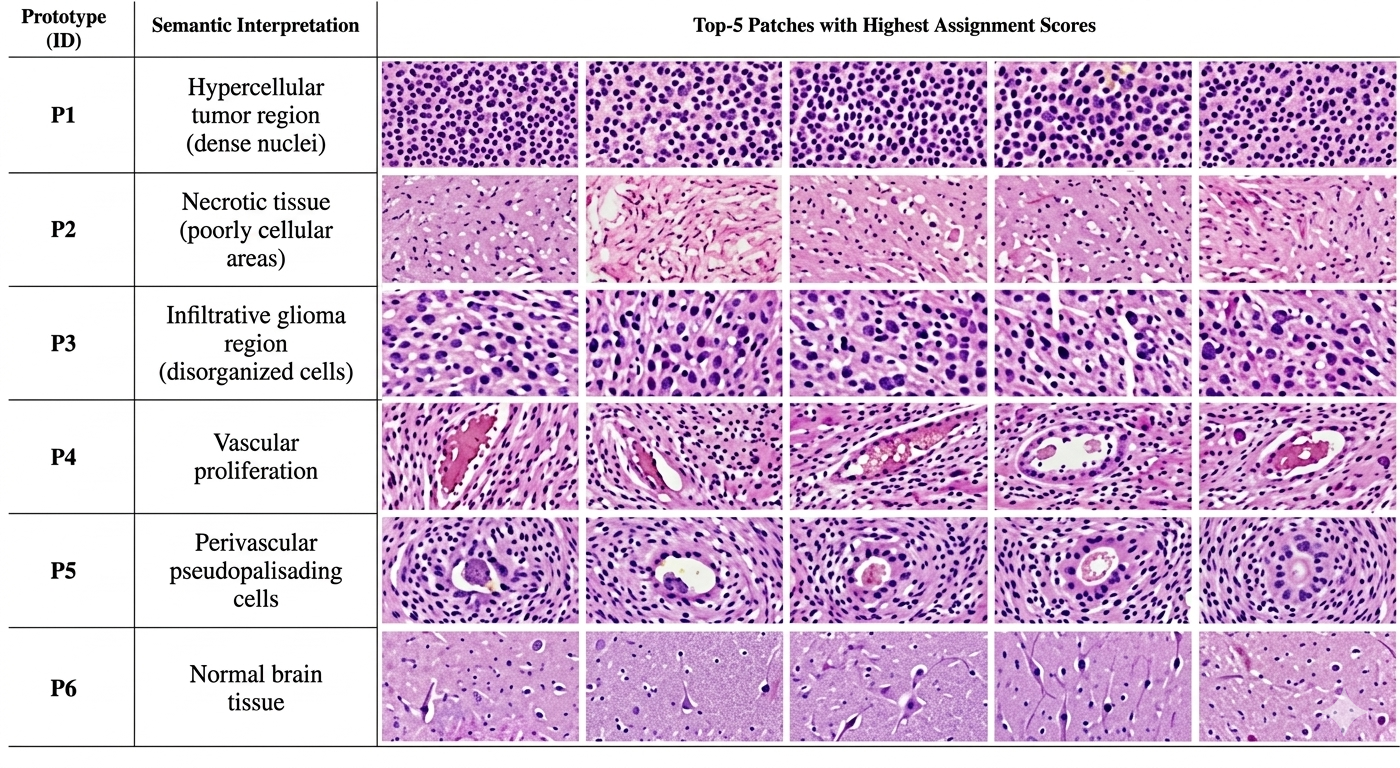}
\caption{Visualization of representative pathology prototypes learned by the proposed PAPD module. For each prototype, the top-5 image patches with the highest assignment probabilities are shown. The learned prototypes capture distinct tissue morphology patterns, demonstrating the interpretability of the proposed pathology-aware representation learning framework.}
\label{fig:prototype_analysis}
\end{figure*}

\subsubsection{Effect of PAPD Loss Weight}
To analyze the influence of the PAPD loss weight, we evaluate the proposed framework using different values of $\lambda$ while keeping all other training settings fixed. As shown in Table~\ref{tab:parameter_experiments}, very small values of $\lambda$ provide insufficient prototype supervision, whereas excessively large values may overemphasize prototype alignment and reduce the contribution of the DINO objective. The best performance is achieved at $\lambda=0.5$, indicating that balanced optimization of self-supervised learning and pathology-aware prototype distillation leads to the most discriminative pathology representations.

\subsection{Prototype Interpretability Analysis}
To better understand the representations learned by the proposed Pathology-Aware Prototype Distillation (PAPD) module, we visualize representative image patches associated with the learned pathology prototypes. For each prototype, the top-ranked patches with the highest prototype assignment probabilities are selected. As shown in Fig.~\ref{fig:prototype_analysis}, patches grouped within the same prototype exhibit similar morphological characteristics, including cellular density, tissue organization, vascular structures, and necrotic regions. The learned prototypes capture distinct pathology patterns that are relevant for glioma characterization, indicating that PAPD organizes tissue representations according to meaningful morphological structures. These observations suggest that the proposed prototype bank provides an interpretable representation space and contributes to the improved classification performance observed in downstream glioma classification tasks.

\subsection{Computational Efficiency Analysis}
CNN-based models offer low computational cost but are limited in capturing long-range contextual information. In contrast, transformer-based architectures generally achieve stronger representation learning at the expense of higher computational complexity. The proposed TVT+PAPD framework maintains a computational cost comparable to standard ViT models while requiring substantially fewer FLOPs than HIPT. Table~\ref{tab:efficiency} shows that the proposed model achieves an effective balance between efficiency and representational capacity, making it suitable for large-scale WSI analysis.

\begin{table}[t]
\caption{Computational Efficiency Comparison of WSI Models}
\label{tab:efficiency}
\small
\renewcommand{\arraystretch}{1.2}
\setlength{\tabcolsep}{2pt}
\centering
\begin{tabular}{lccc}
\hline
\textbf{Model} & \textbf{Params (M)} & \textbf{MACs (G)} & \textbf{FLOPs (G)} \\
\hline
Centralized Baseline (CNN) & 25.56 & 4.13 & 8.27 \\
Standard Vision Transformer (ViT) & 86.24 & 26.86 & 33.73 \\
High Resolution ViT & 86.24 & 35.25 & 48.50 \\
HIPT & 176.32 & 33.73 & 67.40 \\
\hline
\textbf{TVT + PAPD (Ours)} & \textbf{90.00} & \textbf{16.85} & \textbf{33.70} \\
\hline
\end{tabular}
\end{table}

\section{Conclusion}\label{sec:conclusion}
This paper presented TVT-PAPD, a self-supervised framework for whole slide image representation learning that integrates a Tiny Vision Transformer with the proposed Pathology-Aware Prototype Distillation (PAPD) module. By incorporating pathology-aware prototype supervision into teacher-student self-distillation, the proposed method learns more discriminative and interpretable morphological representations from unlabeled histopathology images. Experimental results on the TCGA-LGG/GBM and IPD-Brain datasets demonstrated improved classification performance and robust cross-cohort generalization, while ablation studies confirmed the effectiveness of the proposed PAPD module. Future work will investigate multi-scale representation learning, validation on larger multi-center cohorts, and integration with pathology foundation models for computational pathology applications.

\bibliographystyle{unsrt}
\bibliography{references}

@misc{tcga,
  title = {The {Cancer Genome Atlas} ({TCGA}) {R}esearch {N}etwork},
  howpublished = {\url{https://www.cancer.gov/tcga}},
  note = {Data generated by the TCGA Research Network accessed via the NCI Genomic Data Commons; Last accessed on 2026-07-10.}
}

@article{heath2021nci,
  title = {The {NCI} Genomic Data Commons},
  author = {Heath, Allison P. and Ferretti, Vincent and Agrawal, Sheila and et al.},
  journal = {Nature Genetics},
  year = {2021},
  volume = {53},
  number = {3},
  pages = {257--262},
  doi = {10.1038/s41588-021-00791-5},
}

@misc{gdc,
title= {{Genomic Data Commons (GDC) Data Portal}},
howpublished={\url{https://portal.gdc.cancer.gov/}},
note={Last accessed on 2026-07-10.},
}

@article{litjens2017survey,
  author    = {Litjens, G. and Kooi, T. and Bejnordi, B. and others},
  title     = {A survey on deep learning in medical image analysis},
  journal   = {Med. Image Anal.},
  volume    = {42},
  pages     = {60--88},
  month     = dec,
  year      = {2017},
  doi       = {10.1016/j.media.2017.07.005}
}

@article{campanella2019clinical,
  author    = {Campanella, G. and Hanna, M. G. and Geneslaw, L. and others},
  title     = {Clinical-grade computational pathology using weakly supervised deep learning on whole slide images},
  journal   = {Nat. Med.},
  volume    = {25},
  number    = {8},
  pages     = {1301--1309},
  month     = aug,
  year      = {2019},
  doi       = {10.1038/s41591-019-0508-1}
}

@article{lu2021clam,
  author    = {Lu, M. Y. and Williamson, D. and Chen, T. and others},
  title     = {Data-efficient and weakly supervised computational pathology on whole slide images},
  journal   = {Nat. Biomed. Eng.},
  volume    = {5},
  number    = {6},
  pages     = {555--570},
  month     = jun,
  year      = {2021},
  doi       = {10.1038/s41551-020-00682-w}
}

@article{chen2023gigapath,
  author    = {Chen, R. J. and others},
  title     = {A general-purpose {AI} system for computational pathology},
  journal   = {Nat. Med.},
  volume    = {30},
  number    = {3},
  pages     = {850--862},
  month     = mar,
  year      = {2024},
  doi       = {10.1038/s41591-024-02857-3}
}

@article{lu2024virchow,
  author    = {Lu, M. Y. and others},
  title     = {A multimodal generative {AI} copilot for human pathology},
  journal   = {Nature},
  volume    = {634},
  pages     = {970--978},
  month     = oct,
  year      = {2024},
  doi       = {10.1038/s41586-024-07618-3}
}

@article{wang2022ctranspath,
  author    = {Wang, X. and Yang, S. and Zhang, J. and others},
  title     = {Transformer-based unsupervised contrastive learning for histopathological image classification},
  journal   = {Med. Image Anal.},
  volume    = {81},
  pages     = {102559},
  month     = oct,
  year      = {2022},
  doi       = {10.1016/j.media.2022.102559}
}

@article{zhou2022retccl,
  author    = {Zhou, Y. and others},
  title     = {{RetCCL}: clustering-guided contrastive learning for whole-slide image retrieval},
  journal   = {Med. Image Anal.},
  volume    = {83},
  pages     = {102645},
  month     = jan,
  year      = {2023},
  doi       = {10.1016/j.media.2022.102645}
}

@article{bejnordi2017camelyon,
  author    = {Bejnordi, B. E. and others},
  title     = {Diagnostic assessment of deep learning algorithms for detection of lymph node metastases in women with breast cancer},
  journal   = {JAMA},
  volume    = {318},
  number    = {22},
  pages     = {2199--2210},
  month     = dec,
  year      = {2017},
  doi       = {10.1001/jama.2017.14585}
}

@article{hashimoto2020ai,
  author    = {Hashimoto, D. A. and Rosman, G. and Rus, D. and Meireles, O. R.},
  title     = {Artificial intelligence in surgery: promises and perils},
  journal   = {Nat. Rev. Clin. Oncol.},
  volume    = {17},
  number    = {7},
  pages     = {407--422},
  month     = jul,
  year      = {2020},
  doi       = {10.1038/s41571-020-0341-z}
}

@article{jaume2021explainable,
  author    = {Jaume, G. and others},
  title     = {Quantifying explainers of graph neural networks in computational pathology},
  journal   = {Med. Image Anal.},
  volume    = {70},
  pages     = {102027},
  month     = may,
  year      = {2021},
  doi       = {10.1016/j.media.2021.102027}
}

@article{pinckaers2021detection,
  author    = {Pinckaers, H. and Bulten, W. and van der Laak, J. and Litjens, G.},
  title     = {Detection of prostate cancer in whole-slide images through end-to-end training with image-level labels},
  journal   = {{IEEE} Trans. Med. Imag.},
  volume    = {40},
  number    = {7},
  pages     = {1817--1826},
  month     = jul,
  year      = {2021},
  doi       = {10.1109/TMI.2021.3066295}
}

@article{koohbanani2021selfpath,
  author    = {Koohbanani, N. A. and Unnikrishnan, B. and Khurram, S. A. and Krishnaswamy, P. and Rajpoot, N.},
  title     = {{Self-Path}: self-supervision for classification of pathology images with limited annotations},
  journal   = {{IEEE} Trans. Med. Imag.},
  volume    = {40},
  number    = {10},
  pages     = {2845--2856},
  month     = oct,
  year      = {2021},
  doi       = {10.1109/TMI.2021.3056023}
}

@article{chauhan2024ipdbrain,
  author    = {Chauhan, E. and Sharma, A. and Uppin, M. S. and Kondamadugu, M. and Jawahar, C. V. and Vinod, P. K.},
  title     = {{IPD-Brain}: an {Indian} histopathology dataset for glioma subtype classification},
  journal   = {Sci. Data},
  volume    = {11},
  pages     = {1403},
  month     = dec,
  year      = {2024},
  doi       = {10.1038/s41597-024-04275-z}
}

@inproceedings{dosovitskiy2021vit,
  author    = {Dosovitskiy, A. and Beyer, L. and Kolesnikov, A. and others},
  title     = {An image is worth 16x16 words: transformers for image recognition at scale},
  booktitle = {Int. Conf. Learn. Representations (ICLR)},
  month     = may,
  year      = {2021}
}

@inproceedings{shao2021transmil,
  author    = {Shao, Z. and Bian, H. and Chen, Y. and others},
  title     = {{TransMIL}: transformer based correlated multiple instance learning for whole slide image classification},
  booktitle = {Adv. Neural Inf. Process. Syst. (NeurIPS)},
  volume    = {34},
  pages     = {2136--2147},
  month     = dec,
  year      = {2021}
}

@inproceedings{chen2022hipt,
  author    = {Chen, R. J. and Lu, M. Y. and Williamson, D. and others},
  title     = {Scaling vision transformers to gigapixel images via hierarchical self-supervised learning},
  booktitle = {{IEEE} Conf. Comput. Vis. Pattern Recognit. (CVPR)},
  pages     = {16144--16155},
  month     = jun,
  year      = {2022},
  doi       = {10.1109/CVPR52688.2022.01567}
}

@inproceedings{caron2021dino,
  author    = {Caron, M. and Touvron, H. and Misra, I. and others},
  title     = {Emerging properties in self-supervised vision transformers},
  booktitle = {{IEEE} Int. Conf. Comput. Vis. (ICCV)},
  pages     = {9630--9640},
  month     = oct,
  year      = {2021},
  doi       = {10.1109/ICCV48922.2021.00950}
}

@inproceedings{chen2020simclr,
  author    = {Chen, T. and Kornblith, S. and Norouzi, M. and Hinton, G.},
  title     = {A simple framework for contrastive learning of visual representations},
  booktitle = {Int. Conf. Mach. Learn. (ICML)},
  pages     = {1597--1607},
  month     = jul,
  year      = {2020}
}

@inproceedings{he2020moco,
  author    = {He, K. and Fan, H. and Wu, Y. and Xie, S. and Girshick, R.},
  title     = {Momentum contrast for unsupervised visual representation learning},
  booktitle = {{IEEE} Conf. Comput. Vis. Pattern Recognit. (CVPR)},
  pages     = {9729--9738},
  month     = jun,
  year      = {2020},
  doi       = {10.1109/CVPR42600.2020.00975}
}
\end{document}